# An Overview of the Research on Texture Based Plant Leaf Classification

[1]Vishakha Metre, [2]Jayshree Ghorpade

[1] Computer Engineering Department, Pune University, MITCOE
Pune, Maharashtra 411038, India
*vishakha.metre@gmail.com*

[2] Computer Engineering Department, Pune University, MITCOE
Pune, Maharashtra 411038, India
*jayshree.aj@gmail.com*

**Abstract**
Plant classification has a broad application prospective in agriculture and medicine, and is especially significant to the biology diversity research. As plants are vitally important for environmental protection, it is more important to identify and classify them accurately. Plant leaf classification is a technique where leaf is classified based on its different morphological features. The goal of this paper is to provide an overview of different aspects of texture based plant leaf classification and related things. At last we will be concluding about the efficient method i.e. the method that gives better performance compared to the other methods.

## 1. Introduction

It is well known that plants play a crucial role in preserving earth's ecology and environment by maintaining a healthy atmosphere and providing sustenance and shelter to innumerable insect and animal species [7]. In addition, plant has plenty of use in foodstuff, botany and many other industries [8]. Also World Health Organization estimates that 80% of people in Asia and Africa rely on herbal medicines due to the fact that they are gaining popularity worldwide as they are safe to human health and affordable. Many of them carry significant information for the development of human society [9]. Hence precise identification of the respective plant is vital in treating the patients.

Due to various serious issues like global warming and lack of awareness of plant knowledge, the plant categories are becoming rare and many of them are about to extinct [8]. There is an urgent need for recognizing and classifying plant by its category, to help botanist [8] by setting up a database for plant species.

The objective of this research paper is to concentrate on the plant classification based on the texture of the leaf. Leaf presents several advantages over flowers or fruits in identifying the plant such as its 2-dimensional nature and availability at all seasons worldwide. Moreover texture is the interesting area of research in plant leaf classification filed with newer techniques.

The rest of the paper is organized into the following sections: Section 1 gives an introductory part of the plant leaf classification and its importance in recent years. Section 2 describes a brief literature review on the different texture based plant leaf classification approaches. The analysis of the notion of texture feature is discussed in section 3. Section 4 includes the various popular texture feature extraction methods, followed by section 5 which represents the popular classification techniques in the field of texture. Finally, Section 5 concludes this paper and will be providing future direction.

## 2. Literature Review

Rashad, et al., [3] introduced a novel approach for classification of plants which was based on the characterization of texture properties. They have utilized a combined classifier learning vector quantization along with the radial basis function. The proposed systems ability to classify and recognize a plant from a small part of the leaf is its advantageous thing. Without needing to depend either on the shape of the full leaf or its color features, one can classify a plant having only a portion available that is in itself enough as the proposed system requires only textural features. This system can be useful for the researchers of Botany who need to identify damaged plants, as it can now be done from a small available part. This system is mostly applicable as the combined classifier method produced high performance far superior to other tested methods as its correct recognition rate was 98.7% which has been revealed in the result.

Kadir, et al., [4] proposed a method that incorporates shape, vein, color, and texture features. They have used probabilistic neural networks (PNN) as a classifier for the plant leaf classification. Commonly several methods are there for plant leaf classification but none of them have captured color information, because color was not recognized as an important aspect to the identification. In this case color also playing important role in identification process. The experimental result shows that the proposed method gives average accuracy of 93.75% when it was tested on Flavia dataset which contains 32 kinds of plant leaves.

Sumathi, et al., [5] proposed a feature fusion technique using the Gabor filter in the frequency domain and fusing the obtained features with edge based feature extraction. The extracted features were trained using 10 fold cross validation and tested with CART and RBF classifiers to measure its accuracy. RBF provides a promising accuracy 85.93 % with low relative error for a nine class problem.

Beghin, et al., [6] introduced an approach that combines relatively simple methods which used shape and texture features. The shape-based method extracts the contour signature from every leaf and then calculates the dissimilarities between them. The orientations of edge gradients are used to analyze the macro-texture of the leaf. The results of these methods are then combined with the help of incremental classification algorithm which provides 81.1% accuracy.

Arun, et al., [11] presented an automated system for recognizing the medicinal plant leaves. Texture analyses of the leaf images have been done in this work using the feature computation. The features include grey textures, grey tone spatial dependency matrices (GTSDM) and Local Binary Pattern (LBP) operators. Six different classifiers are used to classify the plant leaves based on feature values. When features are combined without any preprocessing, it resulted into a better classification performance of 94.7%. The dataset comprises of 250 different leaf images, of five species.

The summary of literature review on texture based plant leaf classification is depicted in the Table 1.

## 3. General Classification Approach

Classification process is carried out through number of sub processes. Initially, a leaf image database is constructed which consists of leaf sample pictures with their corresponding plant details. There is a lack of standard leaf image database that can be used for plant classification [6] and therefore, the database is normally constructed by the researchers.

First step for plant leaf classification is image acquisition which includes plucking a leaf and then, capturing the digital image of leaf with digital camera, termed as an input image [6, 7].

In the second step, this image is preprocessed to enhance the important features. This step includes grayscale conversion, image segmentation, binary conversion and image smoothing. The aim of image pre-processing is to improve image data so that it can suppress undesired distortions and enhances the image features that are relevant for further processing [6,7].

Color image of leaf is converted to grayscale image because variety of changes in atmosphere and season cause the color feature having low reliability. Thus it is better to work with grayscale image. Once image is converted to grayscale it is segmented from its background and then converted to binary and performs image smoothing over it [7].

In the next step, the important features are extracted and are matched with the database image. The input image is categorized to the plant whose leaf image has maximum match score using some classifier giving the information of the inputted leaf [6].

The overall classification process is shown in the Fig. 1. Every plant leaf classification technique follows the same process which have been described in this section, only differs in classifier step. Several classification techniques are invented which are chosen depending on the extracted morphological features. Actually, shape, color and texture features are common features involved in several applications. However, some researchers used part of those features only. Vein and contour features are also researcher's interested area of research.

Researchers have used various classification techniques to classify the plants leaves for greater accuracies, considering several morphological features. Currently most of the researchers targeting plant leaf texture as the most important feature in classifying plants.

Table 1: Summary of literature review

| Research Paper | Classification Based On | Classifiers | Features | Advantages | Disadvantages | Accuracy | Dataset Size |
|---|---|---|---|---|---|---|---|
| [1]. Plants Images Classification Based on Textural Features using Combined Classifier [Aug 2011]. | Texture | Combined Classifier (LVQ + RBF) | 1. Ability of classifying and recognizing the plant from small part of the leaf. 2. Useful in cases when plant is damaged etc. | 1. High Performance 2. No need to consider shape or color of leaf. | 1. Do not consider noise. | 98.7 % | 30 |
| [2]. Leaf Classification using shape, color, and texture features [Aug 2011]. | Shape, vein, color, and texture. | Probabilistic neural network (PNN). | 1. Make use of several features for classification. 2. Texture feature is based on lacunarity. 3. Color feature consideration | 1. Better performance of the system due to consideration of several features. 2. Works on large dataset. | 1. Lots of mathematical calculation. | 93.75% | 32 |
| [3]. Edge and texture fusion for plant leaf classification [June 2012]. | Edge and texture. | Radial basis function (RBF). | 1. Extraction of edge & texture feature using Gabor filter and fuse them for image classification. 2. Edge detection by Sobel edge detector for better edge detection with high accuracy. | 1. Use of only two features (edge & texture). 2. Relatively low error with RBF. 3. Works on very large dataset. | 1. Less accuracy compared to other methods. | 85.93% | 132 |
| [4]. Shape and texture based leaf classification [2010]. | Shape and texture. | Incremental classification algorithm. | 1. Fusion of both shape based and texture based analysis. | 1. Combination of relatively simple methods (shape and texture). | 1. Identification of leaves is difficult due to high intra-species and low inter-species variability. | 81.1% | 18 |
| [5]. Texture Feature Extraction for Identification of Medicinal Plants and Comparison of Different Classifiers [Jan 2013]. | Texture [grey textures (i.e. first order), grey tone spatial dependency matrices (GTSDM) and Local Binary Pattern (LBP) operators.] | SGD, kNN, SVM, DT, ET, and RF. | 1. Six different classifiers are used to classify the plant leaves based on feature values. 2. Features are combined without any Preprocessing. | 1. No use of preprocessing increases the classification performance. | 1. As no preprocessing, no noise consideration. | 94.7%. | 250 |

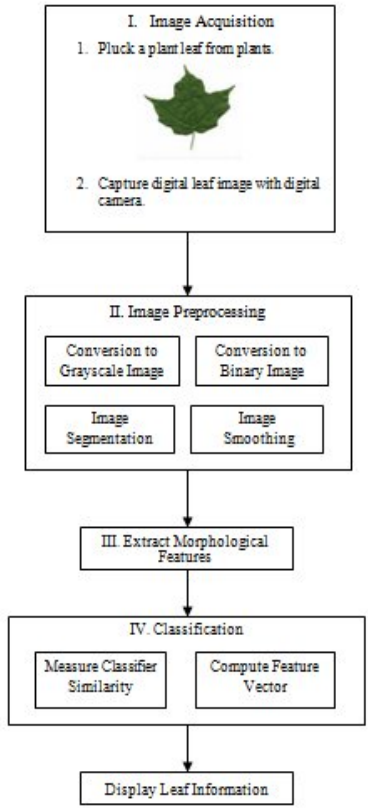

Fig.1 Block diagram for plant leaf classification.

## 4. Analysis of Texture Feature

4.1 Taxonomy of Texture

Texture is important feature considered in an image processing and computer vision field that characterizes the surface and structure of a given object or region. Basically, an image is a combination of pixels and texture is defined as an entity having group of mutually related pixels within an image. This group of pixels is also termed as texture primitives or texture elements (texels) [13].

A texture is usually characterized by the two-dimensional variations in the intensities present in the image. This shows that though there is no precise dentition of texture exists in the literature, but there are a number of intuitive properties of texture which are generally assumed to be true as given below:

i. Texture is a property of areas; the texture of a point is undefined. So, texture is a contextual property and its definition must involve gray values in a spatial neighborhood. The size of this neighborhood depends upon the texture type, or the size of the primitives defining the texture.

ii. Texture constitutes the spatial distribution of gray levels. The two-dimensional histograms or co-occurrence matrices are popular texture analysis tools.

iii. Texture in an image can be alleged at different scales or levels of resolution.

iv. A texture is professed when significant individual "forms" are not present [2].

4.2 Texture Analysis Methods

Textures are a pattern of non-uniform spatial distribution of differing image intensities, which focus mainly on the individual pixels that make up an image. Texture is defined by quantifying the spatial relationship between materials in an image [9]. Image texture has a number of apparent qualities which play an important role in describing texture. Following properties are playing an important role in unfolding texture: uniformity, regularity, density, linearity, directionality, direction, coarseness, roughness, phase and frequency [2].

Seeing that the texture is a quantitative measure of the arrangement of intensities in a region, the methods to characterize texture plunge into four major categories: Statistical, Structural, fractals, and signal processing.

4.2.1 Statistical: Statistical type includes techniques like grey-level histogram, grey-level co-occurrence matrix, auto-correlation features, and run length matrices [2].

i. *First-order texture measures or grey texture* are calculated from the original image values. They do not mull over the relationships with neighborhood pixel. Intensity value concentrations on all or part of an image represented as a histogram is a histogram-based approach to texture analysis. Features resulting from this approach comprise moments for instance mean, standard deviation, average energy, entropy, skewness and kurtosis [4, 13].

ii. *An autocorrelation function* is the measure of the linear spatial relationships between spatial sizes of texture primitives. This approach to texture analysis is rooted in the intensity value concentrations on all or part of an image represented as a feature vector and the calculation of the autocorrelation matrix considers individual pixels [4, 13].

iii. *Run length matrices* exemplify texture images based on run lengths of image gray levels. A run-length matrix p(i, j) is defined as the number of runs with pixels of gray level i and run-length j. Following three matrices describe the traditional run length features: Gray Level Run-Length Pixel Number Matrix (GLRLPNM), Gray-Level Run-Number Vector (GLRNV), and Run-Length Run-Number Vector (RLRNV). These vectors represent the summation of distribution of the number of runs with run length j. However, the original features for run length statistics are Short Run Emphasis (SRE), Long Run Emphasis (LRE), High Gray-Level Run Emphasis (HGRE, and Short Run Low Gray-Level Emphasis (SRLGE) [2].

iv. *A gray level co-occurrence matrix (GLCM)* contains information about the positions of pixels having similar gray level values. It is a two-dimensional array denoted by P consisting of both the rows and the columns signifying a set of possible image values. A GLCM Pd [i, j] is defined by first specifying a displacement vector d = (dx, dy) and counting all pairs of pixels separated by d having gray levels i and j. From this co occurrence matrix, we can derive the following statistics as texture features: Contrast, Dissimilarity, Homogeneity, ASM(Energy), Entropy, GLCM mean, GLCM standard deviation [2,11,13].

4.2.2 Structural: The structural models of texture presume that textures are combinations of texture primitives. The texture is formed by the placement of texture primitives according to certain placement rules. In general, this class of algorithms is restricted in power unless one is dealing with very regular textures. Conceptually, structural texture analysis carried out into two major steps: (a) extraction of the texture elements, and (b) inference of the placement rule. Two different structural methods are considered: two dimensional wavelet transform and Gabor transform [2].

i. *Gabor filters* is a popular signal processing method, which is also known as the Gabor wavelets. The Gabor filters are defined by a few parameters, including the radial center frequency, orientation and standard deviation. The Gabor filters can be used by defining a set of radial center frequencies and orientations which may vary but usually cover 180° in terms of direction to cover all possible orientations [12].

ii. The region-based systems which use *wavelet transform* are classified into three categories: a hierarchical block, a moving window and a pixel. The texture features are calculated from wavelet coefficients of all regions called as subbands. After decomposing the image into non-overlapping subbands, the mean and standard deviation of the decomposed image portions are calculated. The calculated mean and standard deviation represents the texture features for image comparison.

4.2.3 Fractals:

Many natural surfaces possess a statistical quality of roughness and self-similarity at different scales. *Fractals* have become very useful and popular in modeling these properties in the image processing field.

Self-similarity across scales in fractal geometry is a crucial concept. The fractal dimension is the measure of surface roughness. Instinctively saying, larger the fractal dimension, rougher the texture is. Most surfaces are not deterministic although have a statistical variation which makes the computation of fractal dimension more difficult.

The fractal dimension is not sufficient to capture all textural properties. There may be perceptually very different textures that have very similar fractal dimensions. Therefore, another measure, called *lacunarity* has been suggested in order to capture the textural property that will let one distinguish between such textures [4, 13].

4.2.4 Signal processing: Texture is especially suited for this type of analysis because of its properties.

i. *Spatial domain filters* are the most direct way to capture image texture properties.

ii. The frequency analysis of the textured image is best done in the *Fourier domain.* As per the psychophysical results indicated, the human visual system is able to analyze the textured images by decomposing the image into its frequency and orientation components.

iii. *A two-dimensional Gabor function* consists of a sinusoidal plane wave of a certain frequency and orientation modulated by a Gaussian envelope.

## 5. Texture Feature Extraction Methods

The extraction methods are used for extracting interesting and relevant features from the inputted image. The one which is used for the extraction of texture feature from images is called texture feature extraction method. The popular extraction techniques in texture field are discussed in this section.

### 5.1 Grey Level Co-occurrence Matrices (GLCM)

Grey Level Co-occurrence Matrices (GLCM) is a statistical method. It is an old and widely used feature extraction method for texture classification. It has been remained to be an important feature extraction method in the domain of texture classification that computes the relationship between pixel pairs in the image. The textural features can be calculated from the generated GLCMs, e.g. contrast, correlation, energy, entropy and homogeneity. However, in recent years, instead of using the GLCM individually, is combined with other methods. There are a few other implementations of the GLCM, other than the conventional implementation e.g. one-dimensional GLCM, second-order statistical GLCM. It can be also applied on different color space for color co-occurrence matrix [12, 13].

### 5.2 Region Covariance Matrices

The covariance matrix is a common statistical method and is new in the area of texture classification. It is used to calculate the covariance between values. It can also be helpful to generate a covariance matrix from different image features, which are two dimensional matrices with identical sizes generated using edge-based filters. It has fast computations ability because it uses integral images [12, 13].

### 5.3 Gabor Filters

Gabor filters also popular as the Gabor wavelets, is a widely used signal processing method,. The Gabor filters consists of parameters such as the radial center frequency, orientation and standard deviation. It can be can be used by defining a set of radial center frequencies and orientations. Although orientation may vary, it usually covers 180° in direction in order to cover all possible orientations. As signal processing methods produces large feature size, the Gabor filters requires to be downsized for the prevention of the dimensionality issues. Principal Component Analysis (PCA) can be a good choice to downsize the feature space Though Gabor filters are popular in texture classification it sometimes combined with other methods too [12, 13].

### 5.4 Wavelets Transform

Another popularly used signal processing method in image processing and pattern recognition is Wavelet transforms. Currently, it became an important feature to be used in texture classification. Several wavelet transforms are used popularly nowadays such as Discrete Wavelet Transforms (DWT), Haar wavelet and Daubechies wavelets. Among these DWT is most widely used wavelet transform. Similar to the Gabor filters, the wavelet transform are also preformed on the frequency domain rather than the spatial domain of the images. This is because the information on the frequency domain is usually more stable as compared to the spatial domain. Therefore, despite being more complex and slower, wavelet transforms usually produces better features with a higher accuracy [12, 13].

### 5.5 Independent Component Analysis (ICA)

Independent component analysis (ICA) is a computational method for separating a multivariate signal into additive subcomponents supposing the mutual statistical independence of the non-Gaussian source signals. It is a special case of blind source separation. ICA is not very popular method of feature extraction as compared to others, but it can be used to obtained higher order statistics which can be implemented in texture classification. It also overcomes the drawback faced in PCA, i.e. it helps only in obtaining up to second-order statistics [12, 13].

### 5.6 Fractal Measure (Lacunarity)

Other method to get texture features is using fractals. Although, the fractal dimension is not considered for a good texture description, there is a fractal measure known as "lacunarity" which is a measure of non-homogeneity of the data as well as measures lumpiness of the data. It defined in term of the ratio of the variance over the mean value of the function. It may help in distinguishing two fractals with the same fractal dimension. This is because images are not actually fractals, i.e. they do not exhibit the same structure at all scales. Lacunarity is defined by some predefined formulas which were originally applied to grayscale images. But we can also apply them to color images in our implementation using RGB values in order to increase the number of features to represent texture features [4, 13].

## 5.7 Local Binary Patterns (LBP)

Local Binary Pattern (LBP) is a simple and efficient texture extraction method that used to label the pixels of an image by thresholding the neighborhood of each pixel and provides the result as a binary number. This unifying approach is the traditionally divergent statistical and structural models of texture analysis. LBP operator finds its major real-world applications in its robustness to monotonic gray-scale changes caused, such as, by illumination variations. It is possible to analyze images in challenging real-time settings due to its computational simplicity, it. LBP is often used for texture segmentation problems, since it is used to calculate local features [12].

The abstraction of the studied texture feature extraction methods are represented in the Table 2.

Table 2: Texture extraction techniques

| Sr. No. | Techniques | Features | Advantages | Disadvantages |
|---|---|---|---|---|
| [1] | Grey Level Co-occurrence Matrices (GLCM) | 1. It is a tabulation of how often different combinations of pixel brightness values (grey levels) occur in an image. 2. It is usually defined for a series of "second order" texture calculations. | 1. Smaller length of feature vector. 2. Used to estimate image properties related to second-order statistics. 3. It can be improved to be applied on different color space for color co-occurrence matrix. | 1. They require a lot of computation (many matrices to be computed). 2. Features are not invariant to rotation or scale changes in the texture. |
| [2] | Region Covariance Matrices | 1. It is a statistical method used to calculate covariance between two values. 2. Generates two dimensional covariance matrices with identical sizes from different image features. | 1. Low Dimensionality. 2. Scale and illumination independent. 3. Fast computations ability. | 1. Defining a feature mapping vector for RCMs construction is difficult. |
| [3] | Gabor Filters | 1. It is a signal processing method used for defining a set of radial center frequencies and orientations. | 1. It's a multi-scale, multi-resolution filter. 2. It has selectivity for orientation, spectral bandwidth and spatial extent. | 1. Computational cost often high, due to the necessity of using a large bank of filters in most applications. |
| [4] | Wavelets Transform | 1. It is a signal processing method, preformed on the frequency domain of the images rather than the spatial domain. | 1. Produces best features with higher accuracy. | 1. It is more complex and slower. |
| [5] | Independent Component Analysis (ICA) | 1. It decomposes an observed signal (mixed signal) into a set of linearly independent signals. | 1. It is capable of obtaining higher order statistics. 2. It is used to separate a multivariate signal implemented in texture classification. | 1. It is new and not much popular method. |
| [6] | Fractal Measure (Lacunarity) | 1. Lacunarity analysis is a multi-scaled method of determining the texture associated with patterns of spatial dispersion (i.e., habitat types or Species locations) for one-, two-, | 1. It is easily implemented on the computer and provides readily interpretable graphic results. 2. Differences in pattern can be detected even among very sparsely | - |
| | | and three-dimensional data. | occupied maps. | |
| [7] | Local Binary Patterns (LBP) | 1. It labels the pixels of an image by thresholding the neighborhood of each pixel and considers the result as a binary number. | 1. Its robustness to monotonic gray-scale changes caused such as illumination variations. 2. Its computational simplicity. | 1. It is based on the assumption that the local differences of the central pixel and its neighbors are independent of the central pixel itself, & this independence is not warranted in practice. |

## 6. Texture Feature Classification Methods

6.1 k-Nearest Neighbor

k-Nearest Neighbor classifier is used to calculate the minimum distance between the given point and other points to determine which class the given point belongs. It selects the training samples with the closest distance to the query sample. Conceptually, this simple classifier computes the distance from the query sample to every training sample and selects the neighbor or neighbors that are having minimum distance. In terms of plant leaf classification; the distance to be calculated is termed as Euclidian distance. *k-NN* is a popular implementation where *k* number of best neighbors is selected (i.e. k is a small positive integer, k = 1). And the appropriate class is decided based on the highest number of votes from the *k* neighbors [2, 9].

Consider the example shown in Fig. 2 According to the k-NN rule, let k = 7 neighbors of 'W'. Out of seven neighbors, four objects are belonging to class * while three belong to class $.Hence as per the k-NN rule, object 'x' should belong to class *.

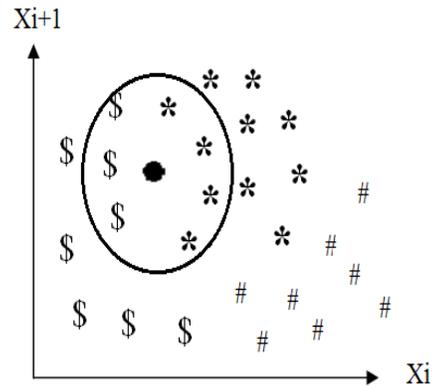

Fig.2 Example showing k-NN classification rule

The nearest neighbor is popular as simpler classifier since it does not include any training process. It is mainly applicable in case of a small dataset which is not trained. However, it suffers the limitation that the speed of computing distance increases according to the number available training samples.

6.2 Learning Vector Quantization

Learning Vector Quantization (LVQ) can be understood as a special case of an artificial neural network, and is a precursor to Self-organizing maps (SOM).It is a supervised version of vector quantization that can be used when we have labeled input data.

An LVQ system can be represented as a set of prototypes given by W= (w(i),..., w(n)) which are defined in the observed data's feature space. According to a given distance for each data point, the prototype that is much closer to the input is measured and the winner prototype is then adapted (i.e. if it is correctly classified it moves closer). If it gets incorrectly classified then moves away.

An advantage of LVQ is that it creates easy to interpret prototypes used by an experts in the respective application domains and also applies to multi-class classification problems yielding variety of practical applications. A key issue in LVQ is the choice of an appropriate measure of distance or similarity for training and classification.

### 6.3 Artificial Neural Networks

ANNs are popular machine learning algorithms that are in a wide use in recent years. Multilayer Perception (MLP) is the basic form of ANN, which is a neural network that updates the weights through back-propagation during the training. Probabilistic Neural Network (PNN) and Convolution Neural Network (CoNN) are the other variations in neural networks, which are recently, became popular in texture classification [10, 12].

a) *Probabilistic Neural Network (PNN)* is derived from Radial Basis Function (RBF) Network and it has parallel distributed processor that has a natural tendency for storing experiential knowledge. It is predominantly a classifier that maps any input pattern to a number of classifications and can be forced into a more general function approximator. A PNN is an implementation of a statistical algorithm called kernel discriminate analysis in which the operations are organized into a multilayered feed forward network having four layers such as Input layer, Pattern layer, Summation layer, and output layer. Fig.3 demonstrates the architecture of PNN classifier considering a general example of BP and Pulse acting as an input vectors [4, 9].

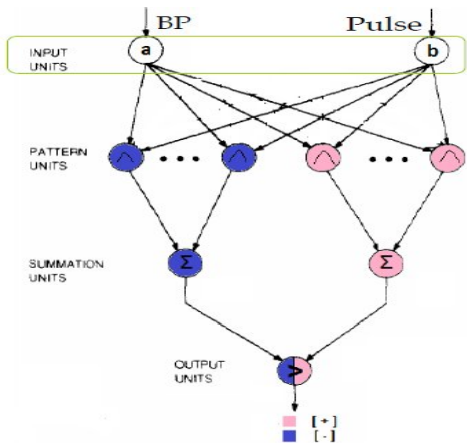

Fig.3 Architecture of probabilistic neural network [7].

b) *Convolutional Neural Network (CoNN)* is a neural network that has convolution input layers that acts as a self learning feature extractor directly from the raw pixels of the input images. Therefore, it can perform both feature extraction and classification under the same architecture [12].

### 6.4 Radial Basis Function

A radial basis function (RBF) is a real-valued function whose value depends only on the distance from the origin. Any function that satisfies this property is a radial function. The frequently used measuring norm is Euclidean distance, but not limited to.(i.e. other distance functions can also be used).

Basically, RBF's are the networks where the activation of hidden units is based on the distance between the input vector and a prototype vector. It has several properties associated with variety of scientific disciplines. This includes function approximation, regularization theory, density estimation and interpolation in the presence of noise. It allows for a straightforward interpretation of the internal representation produced by the hidden layer and training algorithms for RBFs are significantly faster than those for MLPs [2, 3].

### 6.5 Support Vector Machine

Support vector machine (SVM) is a non-linear classifier, which is a newer trend in machine learning algorithm and is popularly used in many pattern recognition problems, including texture classification. In SVM, the input data is non-linearly mapped to linearly separated data in some high dimensional space providing good classification performance. SVM maximizes the marginal distance between different classes. The division of classes is carried out with different kernels.SVM is designed to work with only two classes by determining the hyper plane to divide two classes. This is done by maximizing the margin from the hyper plane to the two classes. The samples closest to the margin that were selected to determine the hyper plane is known as support vectors [9, 11, 12]. Fig.4 shows the support vector machines concept.

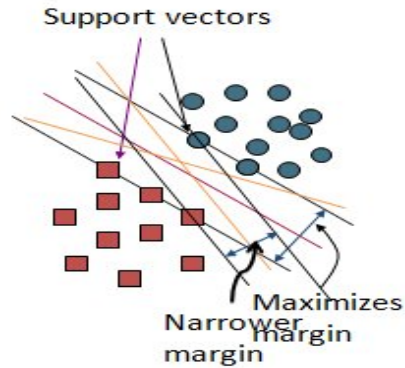

Fig.4 Support vector machine

Multiclass classification is also applicable and is basically built up by various two class SVMs to solve the problem, either by using one-versus-all or one-versus-one. The winning class is then determined by the highest output function or the maximum votes respectively. This leads the multiclass SVM to perform slower than the MLPs.

The main advantage of SVM is its simple geometric interpretation and a sparse solution. Unlike neural networks, the computational complexity of SVMs does not depend on the dimensionality of the input space. One of the drawbacks of the SVM is the large number of support vectors used from the training set to perform classification task. However, SVM is still considered to be powerful classifier, soon to be replacing the ANNs.

6.6 Stochastic Gradient Descent

Stochastic Gradient Descent (SGD) is a simple and efficient approach to discriminative learning of linear classifiers under convex loss functions such as (linear) Support Vector Machines and Logistic Regression. SGD has been successfully applied to large-scale and sparse machine learning problems often encountered in texture classification and natural language processing. If the given data is sparse, then SGD classifiers are efficient to scale the problems having more than $10^5$ training examples as well as more than $10^5$ features [11].

The advantages of Stochastic Gradient Descent are its efficiency and ease of implementation. However the disadvantage of Stochastic Gradient Descent includes its requirement of a number of hyper parameters such as the regularization parameter and the number of iterations along with the sensitivity to feature scaling.

Table 3: Texture Classification Techniques

| Sr. No. | Techniques | Advantages | Disadvantages | Accuracies |
|---|---|---|---|---|
| [1] | k-Nearest Neighbor(k-NN) | 1. Simpler classifier since exclusion of any training process. 2. It is mainly applicable in case of a small dataset which is not trained. | 1. The speed of computing distance increases according to the numbers available in training samples. 2. Expensive testing of each instance. 3. Sensitiveness to noisy or irrelevant inputs. | 85% [3, 9] |
| [2] | Learning Vector Quantization (LVQ) | 1. It creates easy to interpret prototypes. 2. This can be applied to multi-class classification problems and useful in classifying textural features too. | 1. The choice of an appropriate measure of distance or similarity for training and classification. | 98.7 % [1] |
| [3] | Probabilistic Neural Networks(PNN) | 1. Tolerant of noisy inputs and virtually no time consumed to train.. 2. Instances can be classified by more than one output. 3. Adaptive to changing data. | 1. Long training time. 2. Large complexity of network structure. 3. Need lot of memory for training data. | 93.75 % [2] |
| [4] | Radial Basis Function(RBF) | 1. Training phase is faster. 2. The hidden layer is easier to interpret. | 1. When training is finished and it is being used it is slower. So when speed is a factor then it is slower in execution. | 85.93 % [3]. |
| [5] | Support Vector Machine(SVM) | 1. Simple geometric interpretation and a sparse solution. 2. SVMs can be robust, even when the training sample has some bias. | 1. Slow training. 2. Difficult to understand structure of algorithm. 3. Large no. support vectors are needed from the training set to perform classification task. | 92 % [1] |

| [6] | Stochastic Gradient Descent(SGD) | 1. Efficiency. 2. Ease of implementation. | 1. SGD requires a number of hyper parameters such as the regularization parameter and the number of iterations. 2. SGD is sensitive to feature scaling. | 94.7 % [9] |
|---|---|---|---|---|

The various texture classification techniques summarized in Table 3 can be depicted in terms of a bar graph shown in the Fig. 6. The x-axis represents the texture classification techniques while y-axis represents their respective accuracies.

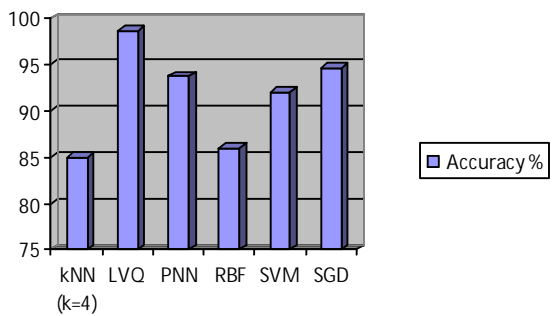

Fig. 6 Plant leaf texture classification techniques with their accuracies.

As per the literature survey and Fig.6, the LVQ method is giving the best accuracy in classifying the plant leaf texture. We can also combine these techniques with other methods in order to achieve a higher accuracy. For example, Rashad, et al., [3] has invented the combined classifier i.e. (LVQ + RBF) giving the maximum accuracy (i.e. (98.7%) in classifying plant leaf texture.

## 7. Conclusion and Future Work

In this survey, we have discussed a brief overview of the notion of Plant classification and its importance in recent years. We have also discussed the different ways in which the problem of accurate plant leaf classification has been formulated in literature [3, 4, 5, 6, 7], and have attempted to provide a taxonomy of texture with its analysis. An overview of the literature on various techniques that can be used for extraction and classification of texture feature are also discussed. Current researches are going on new techniques to be applied for more accurate texture based plant leaf classification.

In our survey, we found that the GLCM and ICA are the new and popular texture extraction methods and combined classifier (LVQ + RBF) is giving the highest accuracy and performance (98.7%) for texture classification [3]. It also present a benefit of identifying the leaf from a small part of the leaf as it uses only texture feature in its classification and hence can be useful for botanist to identify and classify the damaged plant even [3]. Therefore, here we conclude that the combined classifier method proposed in [3] will be the efficient and accurate for texture based plant leaf classification.

As everyone is familiar with the fact that the *"Holy basil"* or commonly known as *"Tulsi"* is also an herbal remedy for a lot of common ailments and it has its own traditional importance in our culture. There may be a case that some plant species having leaves are lookalike *Tulsi.*

Hence, in the future direction, we can work on consideration of noise factor while classification in [3] and can reduce the mathematical operation as much as possible while maintaining the same accuracy or else provide much more accuracy with large databases and use it for the accurate classification of *"Tulsi"* leaves.

## References

[1] A Jiawei Han and Micheline Kamber, "Data Mining Concepts and Techniques," Morgan Kauffman, 2nd Ed, 2006.
[2] Milhran Tuceryan and Anil K. Jian, "Chapter 2.1: Texture Analysis", "The Handbook of Pattern Recognition and Computer Vision" (2$^{nd}$ Edition)(eds),pp.207-248,World Scientific Publishing Co.,1998.
[3] M. Z. Rashad, B. S. el-Desouky,and Manal S. Khawasik, "Plants Images Classification Based on Textural Features using Combined Classifier", International Journal of Computer Science & Information Technology (IJCSIT), Vol 3, No. 4, August 2011,pp.93-100.
[4] Abdul Kadir, Lukito Edi Nugroho, and Paulus Insap Santosa, "Leaf classification using shape, color, and texture", International Journal of Computer Trends & Technology (IJCTT), July-August 2011,pp.225-230.
[5] C. S. Sumathi and A. V. Senthil Kumar, "Edge and Texture Fusion for Plant Leaf Classification", International Journal of Computer Science and Telecommunications, Vol 3, Issue 6, June 2012,pp. 6-9.
[6] T. Beghin, J. S. Cope, P. Remagnino, & S. Barman, "Shape and texture based plant leaf classification", Advanced Concepts for Intelligent Vision Systems (ACVIS),Vol 6475,2010,pp.45-353.
[7] Jyotismita Chaki and Ranjan Parekh, "Designing an Automated System for Plant Leaf Racognition", International Journal of Advances in Engineering & Technology, Vol 2, Issue 1, Jan 2012,pp. 149-158.
[8] N. Valliammal and Dr. S.N Geethalakshmi, "Analysis of the Classification Techniques for Plant Identification through Leaf Recognition" Ciit International Journal of

Data Mining Knowledge Engineering, Vol 1, No. 5, August 2009, pp. 239-243.

[9] Prof. Meeta Kumar, Mrunali Kamble, Shubhada Pawar, Prajakta Patil, Neha Bonde, "Survey on Techniques for Plant Leaf Classification", International Journal of Modern Engineering Research (IJMER), Vol 1, Issue 2,pp-538-544.

[10] H.Fu and Z. Chi, "Combined thresholding and neural network approach for vein pattern extraction from leaf images", IEEE Proc-Vis. Image Signal Process., Vol.153, No 6, December 2006, pp.881-892.

[11] C. H. Arun, W. R. Sam Emmanuel, and D. Christopher Durairaj, "Texture Feature Extraction for Identification of Medicinal Plants and Comparison of Different Classifiers", International Journal of Computer Applications (0975-8887), Vol 62,No.12,January 2013, pp.1-9.

[12] Jing YI Tou, Yong Haur Tav, Phooi Yee Lau, "Recent trends in texture classification: A review", Symposium on Progress in Informaiton & Communication Technology, 2009 pp.63-68.

[13] S. Selvarajah and S. R. Kodiruwakk, "Analysis and Comparison of Texture Based Image Retrieval", International Journal of Latest Trends in Computing, Vol. 2, Issue 1, March 2011,pp. 108-113.

**Vishakha Metre** is a Masters (M.E.) research scholar at the MIT College of Engineering, Pune University, Pune, India. Her research interests include image processing, pattern recognition and data mining.

**Jayshree Ghorpade** biography is working as an Assistant Professor in the department of Computer Engineering in the MIT College of Engineering, Pune University, Pune, India. She has 9.7 years of total experience and her research interests include data structures and networking. Her publications include 3 International journals and she has perceived her Master's degree in Computer Engineering. She been awarded with Best Organized Event Award for TESLA'10 - Codelympics (C/C++ Competition) organized by MITCOE, Pune. She is the author of the book "Computer Networks" published by TechEasy Publications, Pune - March 2011.

.